\documentclass[letterpaper, 10 pt, conference]{ieeeconf}  

\IEEEoverridecommandlockouts                              

\usepackage{graphicx}
\usepackage{subfig}
\usepackage{amsmath}

\usepackage{epsfig} 
\usepackage{mathptmx} 
\usepackage{times} 
\usepackage{amssymb}  
\usepackage{booktabs}
\usepackage{multirow}
\usepackage{algpseudocode}
\usepackage{float}
\usepackage[linesnumbered,noend,ruled]{algorithm2e}
\usepackage{graphicx}
\usepackage{textcomp}
\usepackage{xcolor}
\usepackage{stackengine}

\makeatletter
\renewcommand{\@algocf@capt@plain}{above}
\makeatother

\IncMargin{1em}
\usepackage{hyperref}

\usepackage{comment}
\usepackage{booktabs}
\usepackage{multirow}
\usepackage{algpseudocode}
\usepackage{float}
\usepackage{graphicx}
\usepackage{textcomp}
\usepackage{xcolor}
\usepackage{stackengine}
\usepackage{pgfplots}
\usepackage{pgfplotstable}
\usepackage{ifthen}
\pgfplotsset{compat=1.7}
\usepgfplotslibrary{groupplots}
\usepackage{verbatim}
\usepackage{balance}

\usepackage{pifont}

\newcommand{\red}[1]{\textcolor{black}{#1}}

\newcommand{\cmark}{\ding{51}}%
\newcommand{\xmark}{\ding{55}}%

\title{\LARGE \bf
A Model-Agnostic Approach for Semantically Driven Disambiguation in Human-Robot Interaction
}

\author{Fethiye Irmak Dogan$^{1}$, Maithili Patel$^{2}$, Weiyu Liu$^{2}$, Iolanda Leite$^{1}$, and Sonia Chernova$^{2}$
\thanks{$^{1}$ F. I. Do\u{g}an and I. Leite are with the Division of Robotics, Perception and Learning at KTH Royal Institute of Technology, Stockholm, Sweden
        {\tt\small \{fidogan, iolanda\}@kth.se}}%
\thanks{$^{2}$ M. Patel, W. Liu and S. Chernova are with the School of Interactive Computing at Georgia Institute of Technology, Atlanta, Georgia, USA
        {\tt\small \{wliu88, maithili,  chernova\}@gatech.edu}}%
}

\begin{document}

\maketitle
\thispagestyle{empty}
\pagestyle{empty}

\begin{abstract}

\red{Ambiguities are inevitable in human-robot interaction, especially when a robot follows user instructions in a large, shared space. For example, if a user asks the robot to find an object in a home environment with underspecified instructions, the object could be in multiple locations depending on missing factors. For instance, a bowl might be in the kitchen cabinet or on the dining room table, depending on whether it is clean or dirty, full or empty, and the presence of other objects around it.} 
\red{Previous works on object search have assumed that the queried object is immediately visible to the robot or have predicted object locations using one-shot inferences, which are likely to fail for ambiguous or partially understood instructions.  
This paper focuses on these gaps and presents a novel model-agnostic approach leveraging semantically driven clarifications to enhance the robot's ability to locate queried objects in fewer attempts.  Specifically, we leverage different knowledge embedding models, and when ambiguities arise, we propose an informative clarification method, which follows an iterative prediction process. The user experiment evaluation of our method shows that our approach is applicable to different custom semantic encoders 
as well as LLMs, and informative clarifications improve performances, enabling the robot to locate objects on its first attempts.} The user experiment data is publicly available at 
\href{https://github.com/IrmakDogan/ExpressionDataset}{https://github.com/IrmakDogan/ExpressionDataset}. 

\end{abstract}

\section{Introduction}

\red{Ambiguous instructions are a challenging yet fundamental aspect of human-robot interaction. In a home setting, a user might make requests such as, ``Bring me my cup of coffee" or ``Bring me the cereal bowl". Depending on the context, this could refer to a clean bowl for breakfast or a dirty one while doing the dishes. In such queries, users themselves may be uncertain about the object's location, for example, where they left the cup after having coffee. In these situations, additional semantic information obtained through follow-up questions can provide valuable hints for robots. For instance, knowing that the cup is full of coffee suggests it is unlikely to be in a cabinet, thereby reducing the robot's object search space.}

\red{Prior work has addressed various aspects of handling ambiguous queries, including learning personalized language grounding (e.g., ``my favourite mug") \cite{6281377, churamani2017impact}, disambiguating vague referents in a visual scene (e.g., ``give me that green thing") \cite{dougan2022asking, yi2022incremental, hatori2018interactively, shridhar2020ingress, yang2022interactive, mees2020composing, marge2019miscommunication}, leveraging Large Language Models (LLMs)~\cite{park2023clara, abugurain2024integrating, kobalczyk2025active}, and constructing semantic maps that encode likely object locations \cite{walter2013learning, bastianelli2013line}. In line with this, objects' semantic features have been used to infer missing information in human instructions~\cite{nyga2018grounding, chen2019enabling}, search for objects in homes~\cite{zeng2019generalized, yang2018visual, zheng2022towards}, and manipulate objects based on these properties~\cite{ardon2019learning, WeiyuICRA2020, shridhar2022cliport}. While these studies have made significant progress in resolving ambiguities and modelling object semantics, their predictions have either focused on \textit{fixed visual scenes} and \textit{model-specific disambiguation techniques} 
or relied on \textit{one-shot inferences} without incorporating effective clarification strategies. However, such approaches are likely to fail when a robot operates in a larger space where missing information is crucial for the inference task. Additionally, even with the power of LLMs, generating effective follow-up clarifications 
still poses a challenge~\cite{kuric2024unmoderated}, especially for \textit{what to ask in these questions} in dynamic real-world environments. 
To address these gaps, we propose a novel \textit{\textbf{model-agnostic interactive clarification approach}} that generates informative follow-up questions for a robot's object search in household settings.}

\begin{figure}
    \centering
    \includegraphics[width=0.46\textwidth]{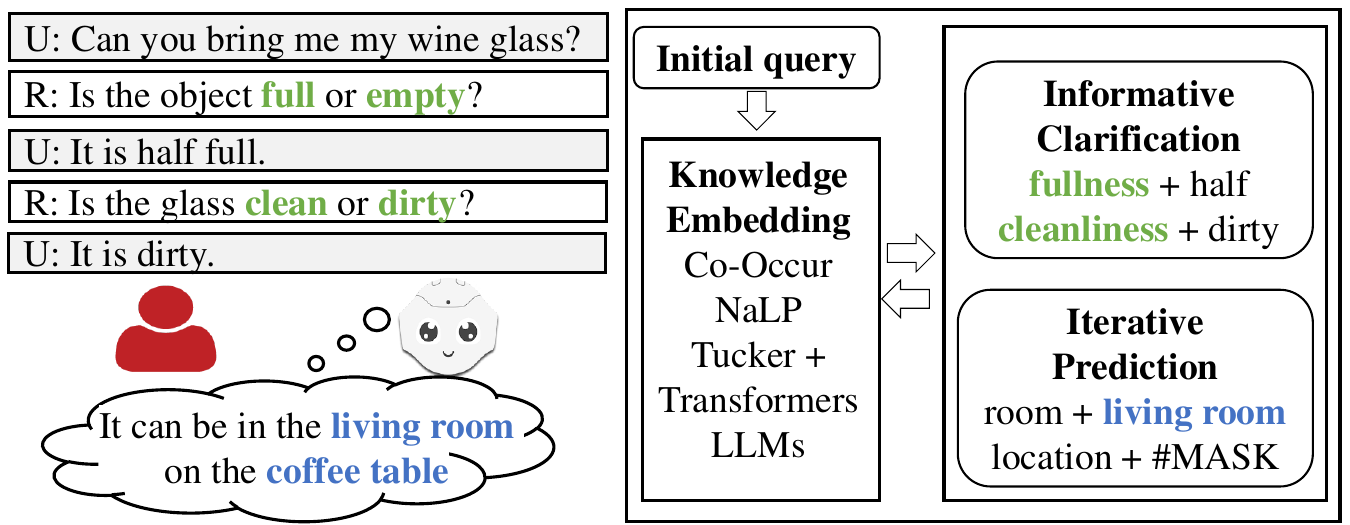}
    \caption{\small{An overview of our method. On the left-hand side, when the user asks the robot to find the wine glass, the robot identifies the input query as ambiguous, then asks for informative semantic properties of the object (in green) and makes the output predictions (in blue) with the gathered new features. To achieve this, on the right-hand side, the initial query is forwarded to the \textbf{knowledge embedding} model to first find the room of the object. Once the initial query is identified as ambiguous, \textbf{informative clarifications} are asked to obtain further knowledge, which is provided to the knowledge embedding model as additional features. After the object's most probable room is inferred, this feature is also given to the embedding, which is then queried to predict the most likely location of the object (\textbf{iterative predictions}).}}
    \label{fig:overview}
    \vspace{-0.7cm}
\end{figure}

Our method enables a robot to successfully fulfil complex, ambiguous commands by eliciting missing information from the user through 
semantically-driven queries that reduce ambiguity. More specifically, these queries are used to guide the robot's object search across an entire household and effectively predict objects' room (e.g., kitchen) and location (e.g., in the sink) properties. To achieve this,  
\red{our method first leverages different knowledge embeddings to receive information about the relationships between given object features and room/location properties}. Then, if there are ambiguities, our interactive system forms informative follow-up questions based on the system's information gain and iteratively infers first the object's room and then its location (see Figure~\ref{fig:overview} for an overview). 
In our evaluations, we conducted an extensive set of experiments. First, we empirically motivate our work by presenting the results of a pre-study in which we collected 80 sample object retrieval instructions given by users to a robot operating in a household setting.  
We found that in 64\% of retrieval instructions, users either did not explicitly state the location of the desired object or provided a location that was incorrect or ambiguous, requiring the robot to infer the correct information.
\red{Then, to evaluate the impact of informative clarifications, we collected a dataset of user object descriptions
through a second user experiment (dataset publicly available). Evaluations on this dataset show that informative clarifications enable the robot to identify the correct room and location of objects within its top predictions consistently across several knowledge embeddings (including LLMs), demonstrating our method's robustness and model-agnostic nature.}

\vspace{-0.2em}
\section{Related Work}
\vspace{-0.5em}
\label{sec:relatedWork}

In complex real-world settings, object disambiguation and follow-up clarifications play a critical role in efficient human-robot interaction (HRI)~\cite{dougan2022asking}. Accordingly, several former studies addressed this challenge via pointing-based or attribute-based questions~\cite{yi2022incremental, shridhar2020ingress,  yang2022interactive, shridhar2018interactive}, using objects' spatial relationships~\cite{dougan2022asking, mees2020composing, pramanick2022doro}, and visualizing the target objects~\cite{hatori2018interactively}. For instance, Yi et al.~\cite{yi2022incremental} have suggested a disambiguation model using scene graphs to identify the described object when there are multiple similar ones in the scene. In another study, Do\u{g}an et al.~\cite{dougan2022asking} have proposed an approach leveraging explainability, and they have generated follow-up clarifications using objects' distinguishing spatial relations. Other studies have approached this problem for object grasping~\cite{yang2022interactive, shridhar2018interactive} or pick-and-place scenarios~\cite{hatori2018interactively, mees2020composing}.

\red{While disambiguating user instructions, recent studies have leveraged the significant potential of Large Language Models (LLMs)~\cite{park2023clara, abugurain2024integrating, kobalczyk2025active}. For instance, Park et al.~\cite{park2023clara} have used LLMs to distinguish ambiguous
or infeasible user requests. More closely related to our work, Jiang et al.~\cite{jiang2024llms} have utilized LLMs for visual object disambiguation through few-shot prompting. While these approaches, including both earlier methods and recent LLM studies, have introduced promising 
ideas for distinguishing similar objects within fixed visual scenes, such as tabletop setups, they have not addressed scenarios where the robot operates in a larger space and the requested object is outside its current field of view. In such cases, asking for follow-up clarifications about an object's semantic features can provide crucial information to resolve uncertainties (e.g., a clean glass cup is more likely to be stored in the cabinet than a dirty glass cup~\cite{Liu-RSS-21}).} 

Semantic understanding of object properties (e.g., a \textit{cup} is \textit{ceramic}, \textit{empty}, located \textit{in kitchen}, and used for \textit{drinking}) has been extensively focused on for improving robot autonomy~\cite{nyga2018grounding, chen2019enabling, zeng2019generalized, yang2018visual, zheng2022towards, ardon2019learning, WeiyuICRA2020, shridhar2022cliport}. Traditionally, large-scale ontologies have been used to encode object knowledge with support for logic and rule-based reasoning~\cite{tenorth2017representations, lemaignan2017artificial}. Due to the ability to infer missing relations, knowledge graph embeddings have been introduced to represent relations between an object's category and its properties (e.g., \textit{mugs} are \textit{ceramic})~\cite{daruna2019robocse,arkin2020multimodal}. 
Additionally, probabilistic logic models have been leveraged to learn task-specific knowledge (e.g., which object shape is most likely to support the pourable affordance)~\cite{ardon2019learning, ChernovaISRR2017,nyga2014pr2, Moldovan2014OccludedOS}. 
To achieve more scalable and fine-grained inference, 
Liu et al.~\cite{Liu-RSS-21} have used Transformer networks while representing inter-connected relations between object properties. 
\red{Further, to improve generalization capabilities, 
Dou et al.~\cite{dou2019domain} learned these features following model-agnostic strategies.} 
Although these models have shown promising results in modelling the semantic properties of objects, they have been based on one-shot predictions, which could fail when there is missing information or ambiguities in user requests. Different from prior methods, instead of one-shot inferences, we use an interactive system that handles such uncertainties with semantically driven follow-up clarifications.

\vspace{-0.1em}
\section{Method}
\label{sec:Method}
\vspace{-0.5em}

\red{For a user-provided object description, we aim for the robot to determine the object's room and location by leveraging clarifications. 
First, we use knowledge embedding models based on custom semantic encoders or LLMs to capture the relations between objects' semantic properties and their location/room features. Next, we introduce an informative clarification technique that maximizes the system's information gain in cases of ambiguity and can be used in conjunction with any of these knowledge embedding models.} 
Finally, we propose an iterative prediction process that first identifies the likely room containing the object and then, conditioned on the room, the object's likely placement.  
An overview of our method is shown in Figure~\ref{fig:overview}.

\vspace{-0.5em}
\subsection{Knowledge Embedding Models}
\label{embedding}
\vspace{-0.15em}

In order to utilize clarifications in the form of semantic object properties, we first use knowledge embeddings to obtain relationships between objects' semantic features and their room/location properties. We represent object properties through feature types $\{t_1, ..,t_n: \forall t_i \in T\}$ and their corresponding values $\{v_1, ..,v_n: \forall v_i \in V\}$ (see Table~\ref{objectFeatures} for a list of properties). The aim of the embedding model is to predict $v_{room}$ and $v_{loc}$ for a given set of object features $\{(t_i,v_i)\}$, obtained from an initial user description $u$ and from follow-up questions $\{(t_m,v_m)\}$. We utilize embedding models using custom semantic encoders and LLMs.


\subsubsection{\red{Custom Semantic Encoders}}
\label{embeddings:custom}
First, we extract object properties $\{(t_i,v_i)\}$ from an object description $u$, e.g., if the description is `the red apple next to the knife', we extract: $\{(class,apple),\ (color,red),\ (reference\ object+knife)\}$. 
Then, we convert each text token into its feature embedding and encode the complete input vector as follows: $h^0 = \{t_1+v_1,\ t_2+v_2, ..,\ t_n+v_n,\ t_{n+1}+\#MASK\}$, where
$t_{n+1}$ (either room or location) is the output feature type whose value is masked and expected to be returned by the embedding. 
After obtaining the input vector, we feed it to one of the following custom semantic encoders:

\textbf{(i) Transformers:}
\vspace{-0.1em}
This model~\cite{Liu-RSS-21} takes the input vector $h_0$ and uses a multi-head attention mechanism, which accounts for the variability of the input's different combinations. To make the predictions, the last feature of the final Transforms layer (which contains the masked feature value) is forwarded to a linear transformation followed by a sigmoid function. This outputs the likelihood of each feature value for the queried feature type. After normalizing the likelihood values, the probability distribution is obtained, which we denote as $p_i \in \mathcal{P}$. 

\textbf{(ii) Co-Occur:} This approach has been used for modelling object co-occurrences~\cite{chao2015mining} and affordances~\cite{8405571} and considers the co-occurrences of object features and their location/room. In other words, this method stores a frequency table that shows how often each object property occurs with each room and location value. In the inference, the probability $p_i$ of each room and location candidate is obtained by normalizing the stored frequencies of given object features. 

\textbf{(iii) NaLP:}
\vspace{-0.1em}
This is a neural network model developed for modelling n-ary relational data in knowledge graphs \cite{guan2019link}.
This model learns to classify whether a combination of object properties, location, and room is valid or not by learning the ``relatedness'' score between these features. To predict $p_i$ for a location or room, we enumerate possible candidate feature values and normalize their corresponding relatedness scores.

\textbf{(iv) Tucker+:}
\vspace{-0.1em}
This model is the extension of TuckER~\cite{balavzevic2019tucker}  suggested by Liu et al.~\cite{Liu-RSS-21} that uses a binary knowledge graph embedding to predict higher-order relations. It is first trained to score binary relations between all possible pairs of input features. To predict an unnormalized score for a candidate room or location at inference time, the binary relations between the candidate and each of the known features are scored and averaged. We then normalize the individual candidate scores to get probabilities $p_i$.


\subsubsection{\red{Large Language Models}}
\vspace{-0.1em}
\red{As an alternative to custom semantic encoders, 
we use language-based sequence prediction models, including the 8B and 70B variants of Meta Llama 3.1 Instruct and GPT 4o and GPT 4o-mini, by providing a single in-context example. Note that we do not pre-process the user description $u$ for these models; instead, we provide the phrase as-is as the target object, and prompt the models to predict the corresponding room or location from an enumerated list of options. 
To predict probabilities $p_i$, we use the next token probabilities normalized across possible options. 
We provide an enumerated list of available options, indexed as A, B, and so on, and rely solely on the token probability of the option index to avoid discrepancies caused by varying token lengths among the options.}

\red{We trained the custom semantic encoders 
on large-scale data (detailed in section~\ref{training}) and used the semantic knowledge inherent in LLMs for zero-shot predictions. In the end, these knowledge sources are utilized to obtain the probability distribution $p_i \in \mathcal{P}$ for room/location predictions. }
After finding the distribution $\mathcal{P}$, we quantify the system's confidence $C$ as the highest probability in $\mathcal{P}$:
\begin{align}
    C \leftarrow \max \{ p_i, \ \  \forall p_i \in \mathcal{P}\}.
\label{eqn:C}
\vspace{-1em}
\end{align}

\noindent If $C$ is higher than a pre-determined threshold $\theta$, 
the output feature value with the highest probability is predicted as the queried feature. On the other hand, if $C \le \theta$, then the query is considered ambiguous, and informative clarifications are generated. As an example, in Figure~\ref{fig:overview}, when the user asks for the `wine glass', the embedding model's confidence in predicting the object's room is low, so the robot needs to ask for follow-up clarifications to reduce its search space.

\setlength{\textfloatsep}{0.5pt}
\begin{algorithm}[t!]

    \caption{The summary of our overall method using custom semantic encoders.
		}
	\label{fig:alg}
	\KwIn{
        user object description ($u$).
        }
         \KwOut{
        object's room ($v_{room}$) and location ($v_{loc}$).
        }
        \vspace{0.5em}\tcc{\textbf{ \ \ \ Knowledge Embedding}}
        \vspace{0.2em}Let $\forall t_i \in T$ be the set of object feature types and $\forall v_i \in V$ be the set of its values\\
        Parse $u$ and obtain $\{t_i+v_i\}_{1}^n$\\
        $h^0 = \{t_i+v_i\}_{1}^n + \{room+ \#MASK\}$\\
        For the given $h^0$, predict the masked value\\
        Let  $\mathcal{P}$ be the prediction's probability distribution\\
        Set the confidence value $C \leftarrow \max \{ p_i, \ \  \forall p_i \in  \mathcal{P}\}$\\
        \vspace{0.5em}\tcc{\textbf{\ \ \ Informative Clarification }}\vspace{0.2em}
        \If{\textnormal{$C \le \theta$ threshold } \textbf{(ambiguous)}}{
        Compute the system entropy $H( \mathcal{P}) = - \sum_i p_i\log p_i$\\
        \For{\textnormal{each $t_i$ in} $T$}{
        Compute $I_{t_i} =  H( \mathcal{P}) - \sum_{v_{t_i}} p^\prime(v_{t_i}|h^0) \cdot H( \mathcal{P}_{t_i}^{v_{t_i}})$\\
        }
        Select $t_m$ with the highest $I_{t_m}$ as the most informative feature\\
        Query user for the value of $t_m$ and set it as $v_{t_m}$\\
        Assign $h^0 = h^0 +  \{t_m+ v_{t_m}\} $\\
        Go back to line 4
        }
        \vspace{0.5em}\tcc{\textbf{\ \ \ \ Iterative Prediction}}\vspace{0.2em}
        \uElseIf{\textnormal{$C > \theta$ threshold } \textbf{(unambiguous)}}{
        \If{\textnormal{masked property type is $room$}}{
        Assign $v_{room}$ as the feature value obtained the probability $C$\\
        Set $h^0 = h^0 + \{room+ v_{room}\} + \{location+ \#MASK\}$\\
        Go back to line 4
        }\vspace{0.3em}
        \uElseIf{\textnormal{masked property type is $location$}}{
        Assign $v_{loc}$ as the feature value obtained the probability $C$\\
        \Return  $v_{room}$ and $v_{loc}$ predictions
        }
        }
\end{algorithm}
\setlength{\textfloatsep}{20pt}

\begin{figure}
    \centering
    \includegraphics[width=0.43\textwidth]{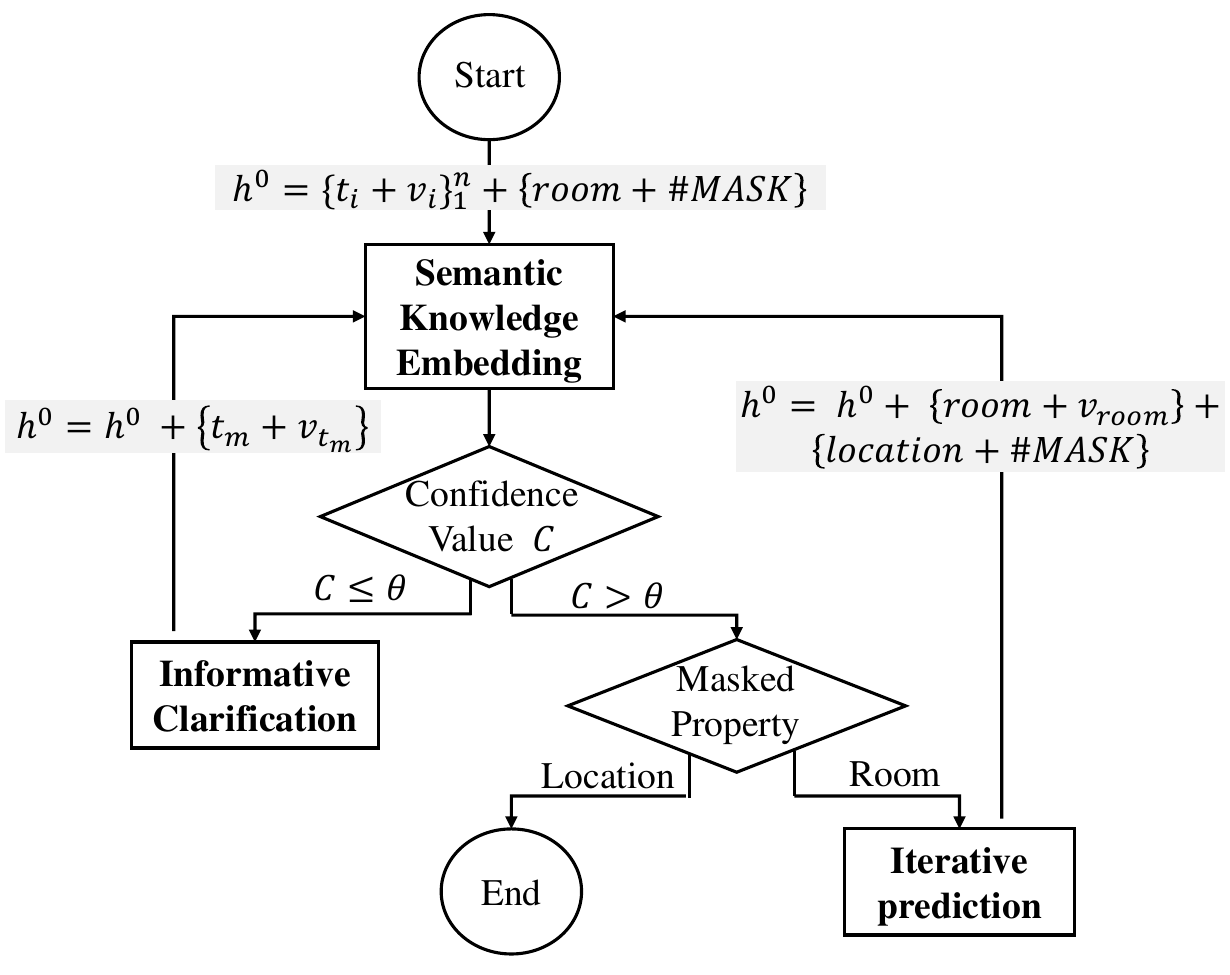}
    \caption{\small{The flowchart summarising our overall approach using custom semantic encoders.}}
    \label{fig:flowchart}
    \vspace{-1.3em}
\end{figure}

\subsection{Informative Clarification}

\red{After obtaining $\mathcal{P}$ from different knowledge embeddings}, when there are uncertainties, we generate an open-ended follow-up question for the feature type that is identified as the most informative. 
To do so, we first measure the system's initial entropy:

\begin{equation}
    H(\mathcal{P}) = - \sum_i p_i\log p_i, \quad \forall p_i \in \mathcal{P}.
\end{equation}

Then, for the custom semantic encoders, we add the candidate feature values of each $t_i \in T$ to the input vector $h^0$. For instance, when  $t_i$ is `fullness', we add $\{fullness+full\}$ and $\{fullness+empty\}$ separately by appending them to $h^0$ as tokenized tuples. On the other hand, for the LLM prompts, we added this information in structured question-answer format (e.g., \textit{``\small\texttt{<Assistant>} Question: fullness \small\texttt{<User>} Answer: full''}). We then compute two new probability distributions $\mathcal{P}_{fullness}^{empty}$ and $\mathcal{P}_{fullness}^{full}$ following the procedure in Section~\ref{embedding}.  Next, we compute the information gain of each $t_i$ by measuring the reduction in the entropy:
\begin{equation}
\vspace{-0.6em}
    I_{t_i} =  H(\mathcal{P}) - \sum_{v_{t_i}} p^\prime(v_{t_i}|h^0) \cdot H(\mathcal{P}_{t_i}^{v_{t_i}}),
\end{equation}
where $I_{t_i}$ is the gain of feature type $t_i$, and $v_{t_i}$ is the candidate values of $t_i$ (e.g., if $t_i$ is `fullness', $v_{t_i}$ is either `full' or `empty'). We assume $p^\prime(v_{t_i}|h_0)$, the conditional probability of observing $v_{t_i}$, come from a uniform distribution. 


After calculating the gain for each $t_i$, the feature type with the highest information gain $t_{m} = argmax(I_{t_i})$ is selected as the most informative, and it is asked in open-ended follow-up clarifications (e.g., `What is the object's material?').  \red{`room' and `location' are assumed to be non-queryable in the follow-up questions, and we also discard the feature types if their corresponding values are already in the initial input vector $h^0$ or in the initial prompt. If no feature results in a positive expected information gain, we discard the clarification and directly predict the room/location.}

\subsection{Inference Phase and Iterative Prediction}
\label{sec:inference}
\vspace{-0.4em}

The inference phase starts with using object properties given in the query $u$ to infer the room value. When there are uncertainties, informative clarifying questions $t_{m} \in T$ are queried, and their responses $v_{t_m} \in V$ are included in the input vector $h_0$ or in the prompt. \red{Once the room value is predicted, this prediction is included in $h_0$ for the custom semantic encoders ($h_0 \leftarrow h_0 + \{room + v_{room}\}$) or included in the prompts for LLMs before inferring the location feature}. 
Then, the process continues to infer the location value, leveraging clarifications when needed (see Algorithm~\ref{fig:alg} and  Figure~\ref{fig:flowchart} for a summary of the procedure).  

\begin{figure}[t!]
\centering
    \includegraphics[width=0.45\textwidth]{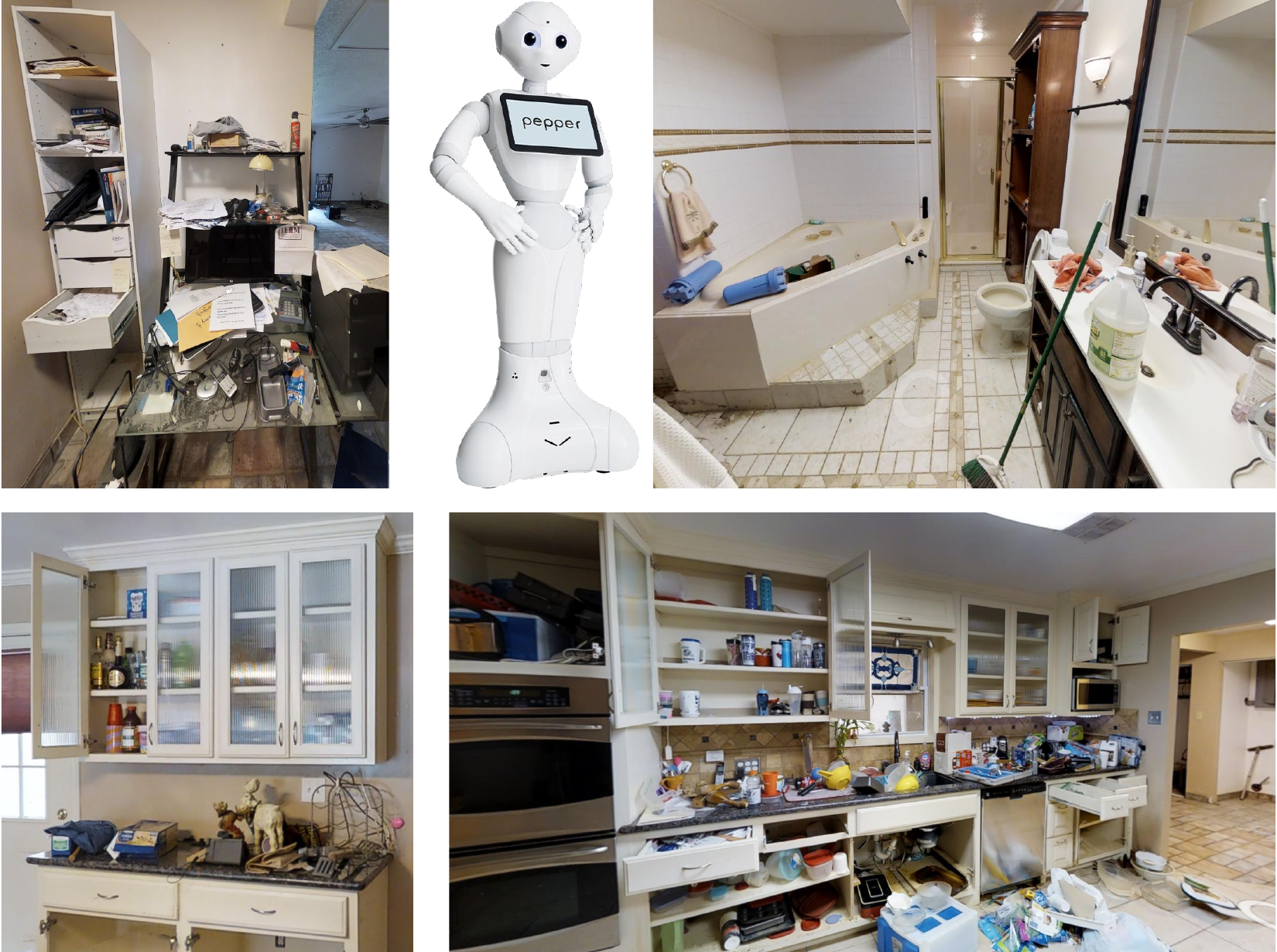}
    \caption{\small{The scenes from a home video clip simulating the robot's object search for given user object descriptions in the pre-study. Given the complexity of the environment, people's descriptions included ambiguities and missing information.}}
    \label{matterport}
    \vspace{-1.5em}
\end{figure}

\section{Pre-Study}
\vspace{-0.4em}
\red{Prior to evaluating our method, 
we conducted a user study to verify our hypothesis that 
human instructions are naturally ambiguous and noisy yet contain diverse semantic features of objects that provide important cues. 
To achieve this, we assessed the correctness and completeness of user instructions, particularly regarding the location of an object, and analyzed the object features most commonly used in descriptions.}

In the pre-study, we asked 10 people, blind to our research question, to watch a home video clip taken from the Matterport dataset~\cite{Matterport3D} containing 3D home scans (example images taken from this home are in Figure~\ref{matterport}). This home contained many self-similar objects in different places (such as containers, bottles, and cups in the kitchen sink, dining room shelves, or bathroom floor). Users first watched a video clip showing a single walk-through of the house.  Then, they were shown images of 8 different objects and, based on the memory of the environment, were asked to instruct a robot to fetch each of the selected items.  Note that due to the clutter in the house, it was impossible for users to memorize the location of all objects; this scenario was designed to simulate a realistic use case in which the user would have forgotten the exact location of an item \red{(e.g., the user is not sure where they left their dirty mug after having coffee and wants the robot to bring it while doing dishes)}.

Using the above process, we collected 80 expressions (10 users and eight objects), which we assessed to determine their ambiguity with respect to describing the object's location.  
We observed that users provided correct room and location information in only 29 out of 80 instructions. 
22 expressions were ambiguous (e.g., the description is `...it's on a shelf' without specifying the room, although there were multiple shelves in different rooms).  In 16 expressions, people specified the place of the objects incorrectly. 
In the remaining 13 expressions, people did not provide any information about the object locations.  These findings support our motivation to design an informative clarification approach eliciting additional information while following user instructions. 


In addition to location properties, we also analyzed the other descriptive object features specified in user instructions to better understand which features users most commonly used to refer to objects. The features shown in the left column of Table~\ref{objectFeatures} were the most commonly utilized ones in user object descriptions (except cleanliness), and accordingly, they were considered in our method to embed them in the semantic knowledge representation. \red{Although cleanliness was not one of the utilized features during the pre-study, we still included this feature in our set as it can provide critical cues in the robot’s
object search across a household. 
It is important to note that not all these features are expected to equally contribute to the robot's object search, but still, they were commonly leveraged in user descriptions, and we opted to include the complete set rather than hand-crafting a selection, as such features have shown to provide meaning information due to the interrelated structure of semantic properties of objects~\cite{Liu-RSS-21}.}

\begin{table}
\caption{The object feature types that are used by people while describing objects in the pre-study and their possible values that are considered in the knowledge embeddings.}
\vspace{-1em}
 \label{objectFeatures}
\begin{center}
\begin{tabular}{ |c|c| } 
 \hline
 Feature Type & Feature Values (Total \# of Values) \\ 
 \hline\hline
 class & clock, laptop, bottle, bowl, fork,\\ 
 & scissors, keyboard, book... (49 values) \\\hline 
 room & bedroom, bathroom, office, kitchen, garage,\\
 & living room, dining room (7 values)  \\\hline 
 location & sink, wall, shelf, counter, floor, kitchen table,\\
 & bedside table, coffee table (8 values) \\\hline
 colour & yellow, brown, green, purple, colourful, pink,\\
 & white, silver, gold... (15 values) \\\hline 
 material & wood, paper, foam, plastic, glass... (7 values) \\\hline 
 reference object & knife, chair, cup, microwave, vase, tv,\\
 & bed, oven, orange, couch... (20 values) \\\hline
 fullness & full, empty, half (3 values)\\\hline
 cleanliness & clean, dirty (2 values)\\
 \hline

\end{tabular}
\end{center}
\vspace{-2.6em}
\end{table}



\section{Analyzing the Knowledge Embeddings}
\label{sec:analysis}
\red{In this section, we provide the training details of the semantic encoders and evaluate the confidence values of these models and LLMs. 
Hence, this section provides our analysis without any clarifications and iterative processes.}

\subsection{Training Custom Semantic Encoders} 
\label{training}
\vspace{-.5em}
To train each of the custom semantic encoders (discussed in Section~\ref{embeddings:custom}) with the features shown in Table~\ref{objectFeatures}, we collected a dataset from Visual Genome (VG)~\cite{krishna2017visual}. VG contains various 2D scenes with their object attributes, object relationships, and scene region descriptions, and these rich features include objects' semantic properties that we aim to learn in the knowledge embedding. We parsed these features and acquired 11,886 different object instances. 70\% of this data was used for training, and the remaining was used for the test and validation (15\% for each). 
While training, the default parameters suggested by~\cite{Liu-RSS-21} were used for each encoder. The models were trained for 100 epochs in Transformers, 10 epochs in NaLP, and 200 epochs in Tucker+ and stored only once in the Co-Occur model.

The HIT scores obtained by each encoder on the VG test set are shown in Table~\ref{semantic_results}, where higher scores show better performances and can range from zero to one. The results show that each semantic encoder successfully learned to model the relationships between object properties and their room and location features, where Transformers obtained the highest scores. 


\subsection{Confidence Values and Thresholds}
\vspace{-0.3em}
To assess the confidence value $C$, we analyzed the correct and incorrect predictions of the test samples using Transformers. We observed that for incorrect room predictions, the encoder's confidence value was 0.47 (STD: 0.13); on the other hand, it was 0.90 (STD: 0.16) for the correct predictions. Similar significant differences were evident for the location predictions (0.79 $\pm$0.15 for correct predictions and 0.56 $\pm$ 0.12 for incorrect ones). This observation backed up our method preferences, i.e., determining when to ask clarifications based on the probability threshold $\theta$, \red{which was experimentally set as 0.65 for the custom semantic encoders. Consistent with prior findings~\cite{wen2024from}, we find LLMs to be over-confident and set the $\theta$ as 0.99.}


\begin{table}
\caption{The HIT scores of custom semantic encoders on the VG test set for the room (R) and location (L) predictions.}
\vspace{-1em}
 \label{semantic_results}
\begin{center}
\begin{tabular}{ |c|c|c|c|c| } 
 \hline
 Encoder Model &R\_H@1 & R\_H@3& L\_H@1 & L\_H@3\\
 \hline
  Transformers & \textbf{0.95} & \textbf{0.99} & \textbf{0.90}  & \textbf{0.97}  \\\hline
 Co-Occur & 0.92 & \textbf{0.99} & 0.57 & 0.83 \\\hline
 NaLP & 0.94  & 0.98  & 0.85 & \textbf{0.97} \\\hline
 Tucker+& 0.91  & 0.98 & 0.68 & 0.82 \\ \hline

\end{tabular}
\end{center}
\vspace{-2.7em}
\end{table}

\section{User Experiment}
After analyzing the knowledge embeddings, we used them in a user experiment to evaluate 
the efficacy of informative clarifications and iterative predictions.


\vspace{-0.3em}

\subsection{Experiment Data}
\vspace{-0.3em}
We first collected video clips of 10 homes from the 3D Matterport dataset~\cite{Matterport3D} (different than the one used in the pre-study). Then, we selected 20 challenging objects from these homes, which might be located in multiple rooms or locations (such as a fork, book, or laptop). It is important to highlight that this dataset didn’t necessarily include multiple similar objects in the same scene, but the ambiguities arise from objects possibly being located in different places. Finally, using the collected home videos and objects, we gathered two sets of data: Expression Data and Object Features Data.

\textbf{Expression Data} was collected from Amazon Mechanical Turk crowd workers to obtain descriptions for the Matterport objects. 40 workers were provided with the home videos (watched one home video at a time and only once), then were shown the object images located at each home. \red{They were asked to describe these objects as if instructing someone/robot to retrieve the displayed object, simulating a robot object search in household setups.}  
In total, we collected 800 expressions (20 objects and 40 users). We processed this dataset by removing expressions such as `I don't know this object' or those consisting of only a single word that names an object other than the provided one.  Further, we discarded expressions if they included objects' rooms or locations, as our aim was for the robot to infer these values. After these, we ended up with 713 expressions. 
\red{The key difference between the Expression Data and the previous dataset~\cite{WeiyuICRA2020} is that the latter was not collected from user object descriptions and primarily consists of a large set of object features. In contrast, the Expression Data allows us to evaluate our system using object features present in user instructions, which inevitably include partial or missing information.}

\textbf{Object Features Data} was collected from the same 20 objects to obtain their features. We asked three human annotators to watch the videos one by one and label the provided objects with the features shown in Table~\ref{objectFeatures}. They could select multiple values for some object features (e.g., colour), and they had the option to indicate `none' or `N/A'. After collecting these features, we checked the cases where three annotators didn't agree with each other. We assumed a feature value applies to an object if the majority selected it. If the selections were on par and did not confirm each other, we set these feature values as `none' or `N/A'. Eventually, we obtained a feature set of these 20 objects that can be queried for any specific value in Table~\ref{objectFeatures}.

\begin{table}
\caption{User experiment results 
for the room (R) and the location (L) predictions. `Inf\_Clar' shows the performance of our approach over the base models.}
\vspace{-1em}
 \label{MTurkResults_embedding}
\begin{center}
\begin{tabular}{ |c|c|c|c|c| } 
 \hline
 Model &R\_H@1 & R\_H@3& L\_H@1 & L\_H@3\\
 \hline\hline
  Transformers~\cite{Liu-RSS-21} & 0.22 & 0.71 & 0.21 &  0.34\\
 Transformers Inf\_Clar & \textbf{0.72} & \textbf{0.90} & \textbf{0.61} & \textbf{0.86}\\\hline\hline
 Co-Occur~\cite{chao2015mining, 8405571} & 0.32 & 0.60 & 0.35 &  0.55\\
 Co-Occur Inf\_Clar & \textbf{0.45} & \textbf{0.89} & \textbf{0.47} & \textbf{0.66}\\\hline\hline
 NaLP~\cite{guan2019link} & 0.28 & 0.73 & 0.25 &  0.42\\
 NaLP Inf\_Clar & \textbf{0.34} & \textbf{0.77} & \textbf{0.36} & \textbf{0.58}\\\hline\hline
 Tucker+~\cite{Liu-RSS-21}& 0.24 &  0.58 & 0.31 &  \textbf{0.60}\\
 Tucker+ Inf\_Clar & \textbf{0.26} & \textbf{0.70} & \textbf{0.35} & 0.56\\
 \hline
 \hline
 Llama 3.1 8B & 0.49 & 0.74 & 0.26 & \textbf{0.54}\\
 Llama 3.1 8B Inf\_Clar & \textbf{0.58} & \textbf{0.79} & \textbf{0.38} & 0.49\\ 
 \hline
 \hline
 Llama 3.1 70B & 0.50 & 0.78 & 0.48 & 0.68 \\
 Llama 3.1 70B Inf\_Clar & \textbf{0.73} & \textbf{0.88} & \textbf{0.64} & \textbf{0.80} \\ 
  \hline
 \hline
 GPT 4o-mini & 0.07  & \textbf{0.67} & 0.14 & 0.43 \\
 GPT 4o-mini Inf\_Clar & \textbf{0.40}  & 0.65 & \textbf{0.18} & \textbf{0.52} \\
  \hline
 \hline
  GPT 4o & 0.46 & 0.83 & 0.31  & 0.58 \\
  GPT 4o Inf\_Clar & \textbf{0.50} & \textbf{0.86} & \textbf{0.45} & \textbf{0.70} \\
 \hline
\end{tabular}
\end{center}
\vspace{-2.5em}
\end{table}
\vspace{-0.5em}
\subsection{Experiment Procedure}
\vspace{-0.5em}
\label{sec:procedure}
\red{First, to preprocess the Expression Dataset for the custom semantic encoders, we parsed the dataset using the spaCy natural language processing library~\footnote{https://pypi.org/project/spacy/}. Each expression was divided into main and relative clauses, from which network inputs were generated by extracting the features listed in Table~\ref{objectFeatures}. On the other hand, for LLMs, user expressions were provided directly in the prompt without any preprocessing.}


When there was a need for disambiguation, the responses to the clarifying questions were obtained by querying the Object Features Dataset. In other words, if an expression is `the metal fork next to the spoon', and the most informative clarifying question is obtained as `Is the object clean or dirty?', the response to this was obtained from the Object Features Dataset and provided as a response. If `none' or `N/A' was received as a response, the question was skipped. The purpose of this approach is to ensure that the responses to the follow-up clarifications are consistent. 
During the experiment, at most two open-ended questions were asked, and after two clarifications, even if the confidence value $C$ was still lower than $\theta$, the inference was made with the gathered information. 
\red{This was to avoid overwhelming users with too many queries during real-time interactions, as suggested in previous studies~\cite{dougan2022asking}.}

\begin{table}
\caption{Performance comparison when follow-up questions are generated by LLMs (`Clar') and our approach (`Inf\_Clar') 
on the user experiment data.}
\vspace{-1em}
 \label{MTurkResults_LLMs}
\begin{center}
\begin{tabular}{ |c|c|c|c|c| } 
 \hline
 Model &R\_H@1 & R\_H@3& L\_H@1 & L\_H@3\\
 \hline\hline
 Llama 3.1 8B Clar & 0.48 & \textbf{0.79} & 0.30 & 0.45\\
 Llama 3.1 8B Inf\_Clar & \textbf{0.58} & \textbf{0.79} & \textbf{0.38} & \textbf{0.49}\\ 
 \hline
 \hline
 Llama 3.1 70B Clar & 0.50 & 0.78 & 0.61 & \textbf{0.84} \\
 Llama 3.1 70B Inf\_Clar & \textbf{0.73} & \textbf{0.88} & \textbf{0.64} & 0.80 \\ 
 \hline
 \hline
 GPT 4o-mini Clar&  0.37 & \textbf{0.67} & 0.14  & 0.44  \\
 GPT 4o-mini Inf\_Clar & \textbf{0.40}  & 0.65 & \textbf{0.18} & \textbf{0.52} \\
  \hline
 \hline
 GPT 4o Clar & \textbf{0.52} & 0.85 & \textbf{0.45} & 0.69 \\
 GPT 4o Inf\_Clar & 0.50 & \textbf{0.86} & \textbf{0.45} & \textbf{0.70} \\
 \hline

\end{tabular}
\end{center}
\vspace{-2.5em}
\end{table}

\vspace{-0.5em}
\subsection{Results}
\vspace{-0.5em}

In this section, we present three sets of results: (i) the impacts of using informative clarifications and iterative predictions with different knowledge embedding models, \red{(ii) the performance of our approach compared to LLMs' own generated clarifications}, and (iii) ablation results showing the contribution of each component. These evaluations were conducted with 713 expressions, and the ground truth was obtained by querying the Object Features Data for the room and location values. Then, HIT scores were calculated by comparing the predictions with the ground truth.

\subsubsection{Different Knowledge Embeddings}
To evaluate the efficiency of our approach, we experimented using different custom semantic encoders and LLMs explained in Section~\ref{embedding}. 
Table~\ref{MTurkResults_embedding} presents the results of these models with and without informative clarifications and iterative predictions. 
\red{The variants using our approach over the base models are marked as `Inf\_Clar'.
The results show that, regardless of the model, employing informative clarifications and iterative predictions improves performance, particularly in HIT@1 scores, which enable the robot to make correct room and location predictions on its first attempt.}

\subsubsection{\red{LLM Clarifications vs Informative Clarifications
}}
\red{In addition to comparing our approach with base language models, we also conducted additional experiments by obtaining variants of LLMs that determine which follow-up questions to ask on their own. (denoted as `Clar' in Table~\ref{MTurkResults_LLMs}). To obtain these variants, we allow LLMs to iteratively predict the room and location while asking up to two questions. 
At each step, LLMs can either ask a question or predict the room or location. If a question is asked, the answer is appended to the prompt for the next iteration. We add the list of all questions and their possible answers in the LLM prompt, thus providing all the information that our informative clarifications have.}

\red{We compare these LLM variants with our informative clarification approach using the user experiment data. The results in Table~\ref{MTurkResults_LLMs} show that when the follow-up clarification was generated by our method (denoted as `Inf\_Clar'), the prediction performances were either on par (for GTP 4o) or better (for the rest of the models, specifically for HIT@1) compared to the LLM-generated clarifications.}

\subsubsection{Ablation Results}


To evaluate the contribution of each component, we conducted an ablation study with the user experiment data. In this evaluation, we used the Transformers networks and assessed the contributions of iterative predictions and informative clarifications.

A non-iterative version is obtained by inferring the room and location values directly from the initial query to evaluate the iterative process. On the other hand, to evaluate the informative clarifications, we obtained a random clarification condition where random open-ended questions were asked when the confidence value $C$ was lower than the threshold $\theta$. 
In both random and informative clarifications, at most two open-ended questions were asked (the average number of questions asked per object was 1.88 and 1.80, respectively).

The results of this evaluation in Table~\ref{MTurkResults} show that both iterative predictions and informative clarifications contribute to the inference process (4th row). When we compare the predictions with and without the iterative process (1st vs 2nd row), we see that the iterative process is helpful for location predictions. Also, the significant performance differences show the importance of clarifying questions (2nd vs 3rd/4th row), and critically, informative clarifications (4th row) boost the performance while identifying the objects' room and location properties. 

\begin{table}
\caption{Ablation results of our suggested method using Transformers semantic encoders.}
\vspace{-1em}
 \label{MTurkResults}
\begin{center}
\begin{tabular}{ c c } 
 \hline
 \multicolumn{2}{c}{Inference Steps}\\
 \hline
 Iterative & Clarification\\
 \hline
 \xmark & \xmark\\
 \cmark & \xmark\\
 \cmark & Random\\
 \cmark & Informative\\
 \hline
\end{tabular}
\begin{tabular}{ c c c c } 
 \hline
 \multicolumn{4}{c}{HIT Scores}\\
 \hline
 R\_H@1 & R\_H@3& L\_H@1 & L\_H@3\\
 \hline
 0.22 & 0.71 & 0.21 &  0.34\\
 0.22 & 0.71 & 0.30 &  0.50\\
 0.58 &  0.84 & 0.50 &  0.74\\
\textbf{0.72} & \textbf{0.90} & \textbf{0.61} & \textbf{0.86}\\
 \hline
\end{tabular}
\end{center}
\vspace{-2.6em}
\end{table}

\section{Discussion, Conclusion and Future Work}
\label{sec:discussion}

\red{In this paper, we present a novel model-agnostic approach that contributed to the robot's object search in household setups. 
First, we obtain knowledge embedding models based on custom semantic encoders or LLMs, and then we leverage these models to form informative follow-up clarifications when there are uncertainties. Finally, the room and location of the queried objects are predicted iteratively. Our evaluations show that our approach is applicable to varying knowledge embedding models, and importantly, informative clarifications are helpful in locating objects on the robot's first attempts.} 

Our pre-study results indicate that ambiguities are very common in user instructions, and finding the object locations based on initial instructions is not always possible. 
After observing the need for object location disambiguation, an initial idea to resolve this issue might be directly asking the object locations in follow-up questions. Under the assumption that users have complete knowledge of the environment, this would be a more efficient solution. \red{However, when people are uncertain or have only a basic knowledge of the environment (as in the pre-study), 
our method can be helpful in reducing object search space and allow the robot find the object locations more efficiently in its top predictions.} An alternative to our approach could be asking people to make logical guesses, 
but this solution (asking people over and over where the object might be) could potentially affect the perceived intelligence and competence of the robot.

\red{The user experiment results in Table~\ref{MTurkResults_embedding} demonstrate that 
follow-up clarifications and iterative processes are helpful regardless of the embedding model, especially for HIT@1 scores. 
This is critical as our approach could be used in small-scale networks (when the inference time is critical) or large-scale models (when accuracy is the priority), and, regardless, it can help the robot improve its object search.
We observed that the custom transformer model shows the largest gain in performance from clarifications, matching or exceeding the LLM performances with clarifications.
We reason that compared to other encoders, Transformers learn the most abstract semantic knowledge, enabling them to benefit the most from clarifications, but still, compared to LLMs, they were able to utilize new information without over-relying on prior knowledge.
Lastly, it is important to note that the custom semantic encoders were not retrained or fine-tuned with the user experiment data, 
and no fine-tuning was done on the language models.
This aims to reflect a real-world scenario where the robot has to use prior knowledge from pretraining and does not have access to a large dataset in the target domain for further fine-tuning. 
As a result of this, the custom semantic encoders' HIT scores were lower in the user experiment than in the VG test set (in Table~\ref{semantic_results}). 
We further reason that this difference is due to the nature of the user experiment: The selected objects (e.g., book, cup, bottle, phone) could appear in multiple rooms or locations, and the user instructions included incomplete information compared to the VG dataset, making the task more challenging.}

\red{The predictions that leverage informative follow-up questions were on par or outperformed LLM-generated clarifications, especially for HIT@1 scores, as shown in Table~\ref{MTurkResults}. Both LLM-generated and informative follow-ups contributed to better predictions compared to the base models (Table~\ref{MTurkResults_embedding}). Still, Llama models received the highest gain from informative clarifications at HIT@1 both for room and location predictions.
The significance of the informative clarifications was also evident from the ablation study  (Table~\ref{MTurkResults}). 
Even random clarifications were helpful for the task, but the highest improvement was achieved through informative follow-ups. This implies that when there are uncertainties, even random information about the specified object could contribute to the robot's object search, which aligns with the previous findings about the interconnected nature of objects' semantics~\cite{Liu-RSS-21}, but informative follow-ups can further enhance interactions.}


\red{Although our primary use case is object search in household settings, future work can adapt our informative follow-up approach to other scenarios where directly requesting missing information is challenging. For example, in elderly care, a robot could infer whether a medication was taken by observing indirect cues, such as a moved pill organizer or an acknowledged reminder, rather than directly asking the user, who may forget or be uncertain. Similarly, given the adaptability of our method across different LLMs, future research can leverage entropy-based clarification techniques for various tasks while utilizing LLMs for categorized predictions. Additionally, our approach can be integrated with visual disambiguation strategies to address the cases where multiple similar objects are present in the same location. This can be achieved by including further visual clarifying questions as suggested by previous studies~\cite{dougan2022asking, yi2022incremental}. Finally, future work can explore the applicability of our approach in environments with different object distributions than typical households, such as home offices and hotel rooms.}

\bibliographystyle{IEEEtran}
\balance{\bibliography{references}}

\end{document}